\documentclass[fleqn,10pt]{wlscirep}
\usepackage[utf8]{inputenc}
\usepackage[T1]{fontenc}
\title{The Governance of Physical Artificial Intelligence}

\author[1,*]{Yingbo Li*}
\author[2]{Anamaria-Beatrice Spulber}
\author[3,*]{Yucong Duan*}
\affil[1]{Hainan University}
\affil[2]{Visionogy}
\affil[3]{Hainan University}
\affil[*]{Corresponding author: xslwen@outlook.com, duanyucong@hotmail.com}

\begin{abstract}
Physical artificial intelligence can prove to be one of the most important challenges of the artificial intelligence. The governance of physical artificial intelligence would define its responsible intelligent application in the society.

\end{abstract}
\begin{document}

\flushbottom
\maketitle
%
%


\section*{Introduction}
Artificial Intelligence (AI) has grown to be the fundamental technology in today's world. Over the last few years, not only has AI been popularly applied in typical AI applications of information and signal processing such as Natural Language Processing (NLP), but it has also empowered all the other industries such as healthcare and robotics. Miriyev and Kovac \cite{pai} proposed to define the AI used in Robotics as Physical Artificial Intelligence (PAI) because PAI interacts with the physical world, contrary to the notion of traditional Digital Artificial Intelligence (DAI) applied in digital information processing. From this perspective, we propose to extend the notion of PAI to a much wider domain to also include Internet of Things (IoT), or automatic driving cars. To the best of our knowledge, most research on AI governance is limited to the domain of DAI, so in the present paper we propose to outline the governance framework of PAI.

\section*{The application of PAI}

\begin{figure*}[ht]
\centering
\includegraphics[width=0.9\columnwidth]{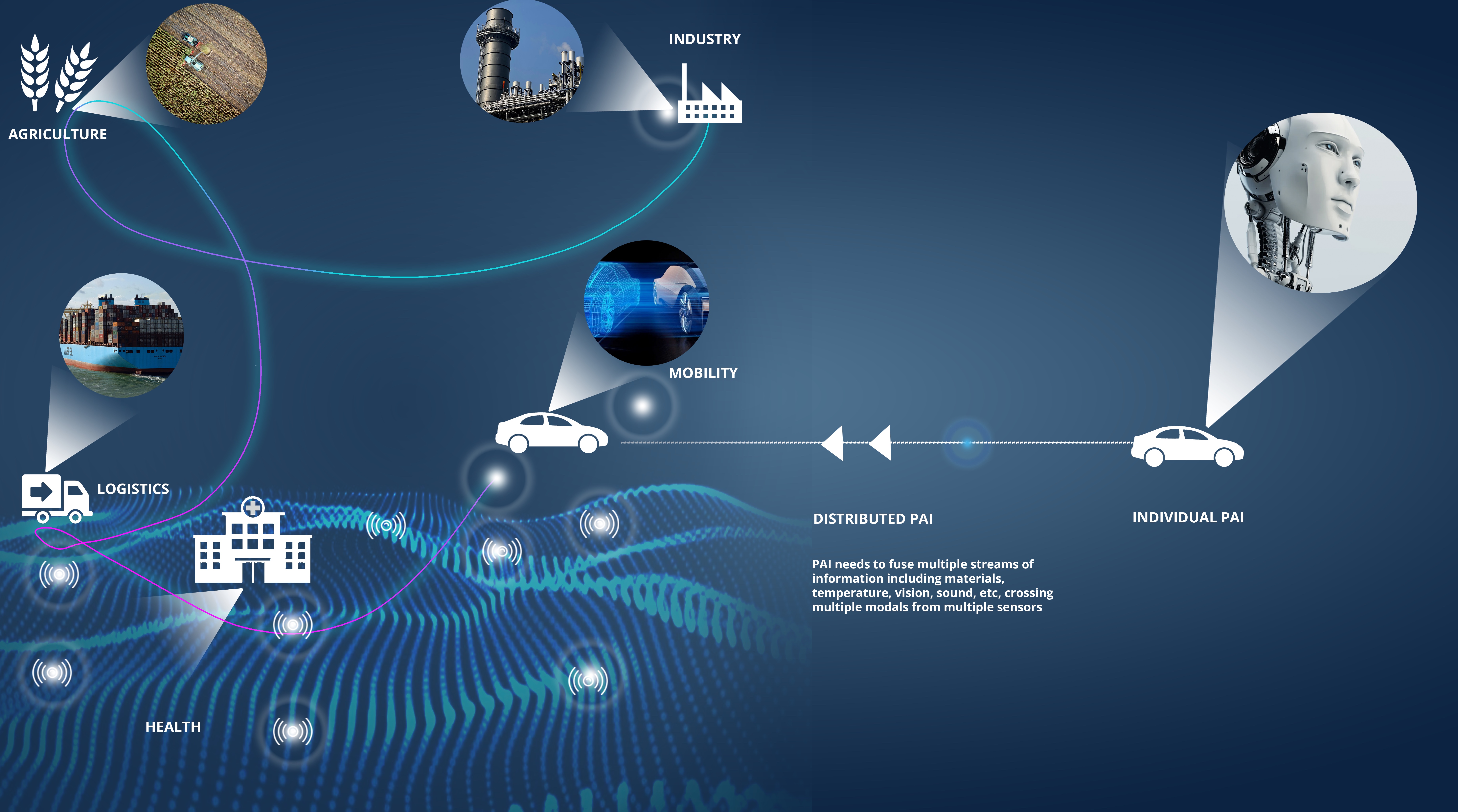}
\caption{The applications of Distributed PAI}
\label{fig:dpaiin}
\end{figure*}

In the proposed concept of PAI by Miriyev and Mirko~ \cite{pai}, PAI refers to the typical robot system. While, we propose to extend the concept of PAI to cover all potential applications with the built-in AI perceiving and interacting between the cyberspace and the physical world. Besides the robot system with the AI working in an integrated and limit physical environment, the distributed intelligent system with AI capability is the typical Distributed PAI. As shown in Fig. \ref{fig:dpaiin}, PAI could be applied in and include multiple distributed industries, such as IoT, self-driving cars, agriculture, healthcare and logistics.

We propose to classify PAI into two overlapped kinds as shown in Fig. \ref{fig:dpaiin}: Independent PAI and Distributed PAI. Independent PAI refers to the intelligent device and the robot~\cite{pai}. Distributed PAI becomes more and more popular when the edge computing~\cite{edge} is mature and every device is connected to the network in the wider space. IoT and edge computing are typical Distributed PAI subdomains. Since it is popular for every intelligent system to be online and individual units in Distributed PAI have strong computing capabilities now, Independent PAI and Distributed PAI will overlap in multiple applications \cite{aip}.

\begin{figure*}[ht]
\centering
\includegraphics[width=0.6\columnwidth]{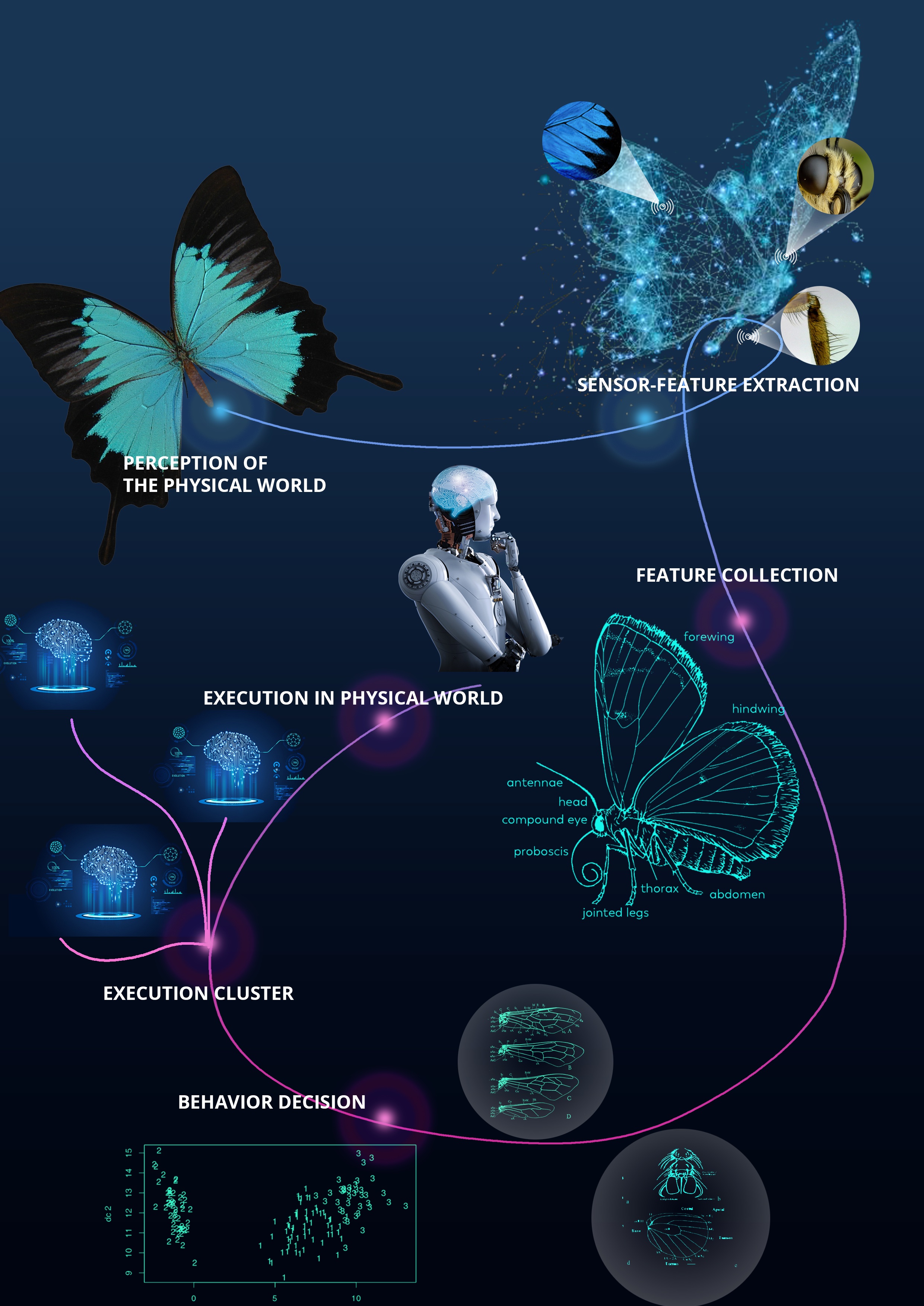}
\caption{Distributed PAI}
\label{fig:dpai}
\end{figure*}

The IoT is a typical distributed system with a spatial distribution that ranges from a small space such as a room to a wider area such as a city. The IoT is formed of various sensors that capture the signals and changes in the physical world. Its AI power could happen in both server side and the edge side. Based on the AI analysis, IoT could directly or indirectly make predictions in the cyberspace to influence the physical world. For example, a self-driving car needs to first perceive real time road situations and connect to the Internet for navigation, then adjust the driving behavior. The agriculture is one of the most successful PAI applications. The sensors in the agriculture including cameras, temperature meter, hygrometer, etc, monitor the growth of the plant and predict, for example, the optimal pesticide intervention and the best harvesting time. In the healthcare industry, families, nursing homes or hospitals could use the biological sensors and the chemical sensors to monitor a patient and predict potential risks such as falling or unstable situations through a monitoring center. The "last mile" is the expensive and hard problem in the logistic industry. A typical distributed AI application could help with delivery tasks through delivery robots and drones connected to and commanded by the center server. Another example is the automatic sorting robot that has been used in the sorting center of the logistics. The general framework of Distributed PAI is described in Fig. ~\ref{fig:dpai}.

DAI mimics the brain capability of logical thinking and induction in human brain, to process the data and signals perceived by human eyes and ears. The human brain is only responsible for processing the signals and transmitting commands to other parts of the body such as movement, vision perception, sound perception, digestion and etc. By comparison, Individual PAI is like an individual human body, while Distributed PAI further extends the AI capabilities just like the human society is composed of multiple humans.

Due to the probabilistic data processing of the current DAI, decisions from DAI are unsubstantial, not being able to reduce the uncertainty of applications. On the other side of promoting the application of DAI, the explainable property of current DAI poses increasing governance challenges against negative and malicious practice of DAI, including data biases and AI frauds, etc. Explainable AI tries to explain and understand the internal operating mechanism of AI. Distributed PAI interacts with the physical world with a much larger spatial area and consequently accumulates Big Data AI footprints which includes much longer interaction trajectories crossing cyberspace and physical world. So explainable AI applied in Distributed PAI has the advantage to reveal the internal mechanism of DAI and the integrated human-cyber-physical social phenomena. 


PAI needs to combine multiple streams of information including materials, temperature, vision, sound, etc, crossing multiple modals from multiple sensors as shown in  Fig.~\ref{fig:dpaiin}. Through the mix of the multimodal information, PAI builds competitive capability that uses multiple types of information which allows it to make better decision and better precision, in the context of imprecise data collection, inconsistent information and incomplete knowledge scattering over various abstraction levels. The various sources of data and information bring multiple kinds of data, which outperform a single source of data, to make real-time decisions and predictions. This is a significant feature of PAI.

\section*{The governance of PAI}

The governance of DAI has been challenged by researchers from a fairness to social impact perspective. DAI has been facing the challenges of risk and governance \cite{gov} from, but not limited to, the following aspects: 1) The storage and transfer security; 2) The fake data; 3) The social privacy; 4) The bias of the sex, gender, and race because of the limited training datasets.

As a consequence to the multi-source perception and multi-dimension interaction in a much larger space, PAI, especially Distributed PAI, brings more uncertainty and risk from the social impact to the technology influence:
\begin{itemize}
  \item The existence problem. PAI especially Distributed PAI such as IoT needs multiple kinds of sensors to interact with the physical world. If PAI is distributed in a limited space such as a factory, it will not encounter challenging regulation problems because it is in an internal space. However, if the space is extended to a larger space such as a city which is not under one unique regulation, PAI will face problems of social regulations.
  \item The data organization problem. The multiple sources of data from the physical world of a wider space will increase the structural construction and integration complexity of the data and information. The Knowledge Graph could be the potential solution for the information organization in the hierarchical structure.
  \item Cannikin Law. The development of PAI depends on at least 5 disciplines of materials science, mechanical engineering, chemistry, biology and computer science. Therefore, the slower development of one discipline will cause the problem of Cannikin law and prohibit the development of PAI.
  \item The social acceptance. Similar to the dilemma of DAI, the ubiquitous application of PAI will cause the worry of the society regarding the increase of unemployment, broadening of the gap in income, the shrinking of privacy space, etc. The acceptance from multiple aspects of the law and society will influence the application of PAI in the research, the industry and the society. So, the acceptance of the society and the corresponding legislation is a potential factor for PAI.
\end{itemize}

\begin{figure}[ht]
\centering
\includegraphics[width=0.7\columnwidth]{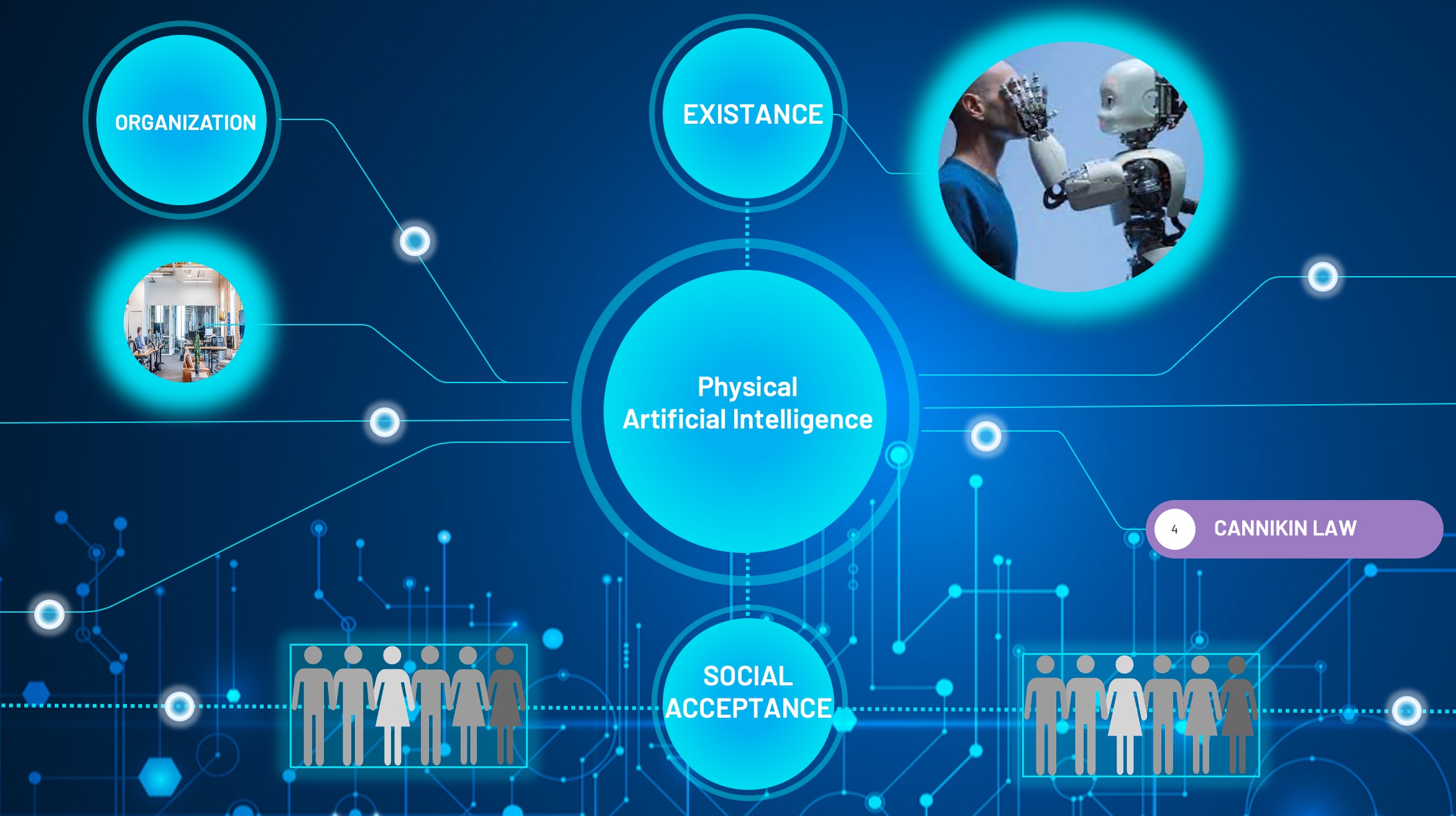}
\caption{PAI governance problems}
\label{fig:dikwpai}
\end{figure}

We illustrates above governance problems of PAI in Figure \ref{fig:dikwpai}. The development of PAI has to resolve these four problems.

\section*{Conclusion}
We have suggested extending the notion of PAI to a larger physical space with the distributed applications with the notion of Distributed PAI. The spatial variety of Distributed PAI could vary from a room space to a city space, while its applications include IoT, agriculture, and so forth. Since PAI, especially Distributed PAI, perceives and interacts with different physical entities in different spaces, the governance issues including the existence problem has been challenged. We have put forward a framework of governance problems of PAI however this is open to further discussions since it is a research topic applying not only to the research but also the development of the whole human society.


\begin{thebibliography}{00}

\bibitem{pai} Miriyev, Aslan, and Mirko Kovač. "Skills for physical artificial intelligence." Nature Machine Intelligence 2.11 (2020): 658-660.
\bibitem{edge} Satyanarayanan, Mahadev. "How we created edge computing." Nature Electronics 2.1 (2019): 42-42.
\bibitem{aip} Vinuesa, Ricardo, et al. "The role of artificial intelligence in achieving the Sustainable Development Goals." Nature communications 11.1 (2020): 1-10.
\bibitem{gov} Margetts, Helen, and Cosmina Dorobantu. "Rethink government with AI." Nature (2019): 163-165.















\end{thebibliography}
\end{document}